\def\year{2020}\relax
\documentclass[letterpaper]{article}

\usepackage{aaai20}
\usepackage{times}
\usepackage{helvet}
\usepackage{courier}
\usepackage{graphicx}
\usepackage{subcaption}
\usepackage{comment}
\usepackage{amssymb}
\usepackage[hyphens]{url}
\usepackage[linesnumbered,vlined,ruled]{algorithm2e}
\usepackage{fancyhdr}

\SetKwFunction{AvgToFroDist}{Average-Distance}
\SetKwFunction{FastMapD}{FastMap-D}
\SetKwProg{function}{Function}{}{}
\SetKwProg{sub}{\(\bullet\)}{}{}

\makeatletter
\newcommand*{\centerfloat}{
\parindent \z@
\leftskip \z@ \@plus 1fil \@minus \textwidth
\rightskip\leftskip
\parfillskip \z@skip}
\makeatother

\title{Embedding Directed Graphs in Potential Fields Using FastMap-D}
\author{Sriram Gopalakrishnan$^1$ \and Liron Cohen$^2$ \and Sven Koenig$^2$ \and T.~K.~Satish Kumar$^2$ \\
$^1$Arizona State University \\
$^2$University of Southern California \\
sgopal28@asu.edu, \{lironcoh, skoenig\}@usc.edu, tkskwork@gmail.com}

\begin{document}
\maketitle

\begin{abstract}

Embedding undirected graphs in a Euclidean space has many computational benefits. FastMap is an efficient embedding algorithm that facilitates a geometric interpretation of problems posed on undirected graphs. However, Euclidean distances are inherently symmetric and, thus, Euclidean embeddings cannot be used for directed graphs. In this paper, we present FastMap-D, an efficient generalization of FastMap to directed graphs. FastMap-D embeds vertices using a potential field to capture the asymmetry between the pairwise distances in directed graphs. FastMap-D learns a potential function to define the potential field using a machine learning module. In experiments on various kinds of directed graphs, we demonstrate the advantage of FastMap-D over other approaches.

\end{abstract}

\section{Introduction}

Graph embeddings have been studied in multiple research communities. For example, in Artificial Intelligence (AI), they are used for shortest path computations~\cite{cohen2018fastmap} and solving multi-agent meeting problems~\cite{jiaoyang2018}. In Knowledge Graphs, they are used for entity resolution~\cite{bordes2013}; and in Social Network Analysis, they are used for encoding community structures~\cite{perozzi2014socialnets}. In general, graph embeddings are useful because they facilitate geometric interpretations and algebraic manipulations in vector spaces. Such manipulations in turn yield interpretable results in the original problem domain, such as in question-answering systems~\cite{bordes2014}.

Despite the existence of many graph embedding techniques, there are only a few that work in linear or near-linear time\footnote{linear time after ignoring logarithmic factors}. FastMap~\cite{cohen2018fastmap,jiaoyang2018} is a near-linear-time algorithm that embeds undirected graphs in a Euclidean space with a user-specified number of dimensions. The efficiency of FastMap makes it applicable to very large graphs and to dynamic graphs such as traffic networks and marine environments for unmanned surface vehicles.

The resulting Euclidean embedding can be used in a variety of contexts. The $L_1$-variant of FastMap~\cite{cohen2018fastmap} produces an embedding useful for shortest path computations. Here, the $L_1$ distances between the points corresponding to pairs of vertices are used as heuristic distances between them in the original graph; and this $L_1$ distance function is provably admissible and consistent, thereby enabling A* to produce optimal solutions without re-expansions. The $L_2$-variant of FastMap~\cite{jiaoyang2018} produces an embedding that is generally useful for geometric interpretations. In the multi-agent meeting problem, for example, the problem is first analytically solved in the Euclidean space and then projected back to the original graph using Locality Sensitive Hashing (LSH)~\cite{datar2004locality}.

In general, the properties of the Euclidean space can be leveraged in many ways. For example, a Euclidean space is a metric space in which the triangle inequality holds for distances. In addition, in a Euclidean space, geometric objects, like straight lines, angles and bisectors, are well defined. The ability to conceptualize these objects facilitates visual intuition and can help in the design of efficient algorithms for Euclidean interpretations of graph problems.

Despite the usefulness of Euclidean embeddings, Euclidean distances are inherently \emph{symmetric} and, thus, cannot be used for \emph{directed} graphs. Directed graphs arise in many real-world applications where the relations between entities are asymmetric, such as in temporal networks and social networks. In this paper, we present FastMap-D, an efficient generalization of FastMap to directed graphs. FastMap-D embeds vertices using a \emph{potential field} to capture the asymmetry between the pairwise distances in directed graphs. Like the $L_2$-variant of FastMap, FastMap-D focuses on minimizing the distortion between pairwise distances in the potential field and the corresponding true distances in the directed graph. FastMap-D therefore provides physical interpretations of problems posed on directed graphs by enabling vector arithmetic in potential fields. It is a back-end algorithm meant to support a number of applications, including question-answering, machine learning and multi-agent tasks on directed graphs.\footnote{Shortest path computation is just one more application that requires FastMap-D to also consider the properties of admissibility and consistency if the intended search framework is related to A*.}

The difference in potential of two points in a potential field is inherently asymmetric and therefore a good choice for capturing distances in a directed graph. FastMap-D constructs a potential function defining the potential field using a machine learning module. Through experiments conducted on various kinds of directed graphs, we demonstrate the advantage of FastMap-D over other approaches, including those that directly apply machine learning techniques to learning pairwise distances between the vertices.

\section{Related Work}

There are a variety of approaches that embed \emph{undirected} graphs in a Euclidean space. While some approaches simply try to preserve the pairwise distances between vertices in the embedding, other approaches try to meet additional constraints. For example,~\cite{linial1995geometry} surveys several methods for the low-distortion embedding of undirected graphs and their usage in algorithmic applications such as clustering. On the other hand, the Euclidean Heuristic Optimization (EHO)~\cite{RBS:AAAI:11} and the $L_1$-variant of FastMap~\cite{cohen2018fastmap} meet additional constraints on the pairwise distances in the embedding. In particular, these distances satisfy admissibility and consistency, which are useful for shortest path computations with heuristic search.

Global Network Positioning~\cite{ng2002predicting} first embeds landmarks in a Euclidean space and then uses them for a frame of reference. Like EHO, this algorithm relies on solving Semi-Definite Programs (SDPs) or similar approaches that are prohibitively expensive for large graphs. Big-Bang Simulation (BBS)~\cite{shavitt2004big} is a different method that simulates an explosion of particles under a force field derived from the embedding error. Although it does not rely on solving SDPs, it is still prohibitively expensive for large graphs.

Existing works on embedding \emph{directed} graphs, such as Node2Vec~\cite{node2vec}, LINE~\cite{LINE} and APP~\cite{APP}, focus on preserving the \emph{proximity} of vertices. First-order proximity refers to the distance between two vertices; second-order proximity is related to the similarity of their one-hop neighboring vertices; third-order proximity is related to the similarity of their two-hop neighboring vertices; and so forth. These proximity-preserving embedding algorithms are based on skip-gram models originally developed in the context of Natural Language Processing~\cite{word2vec}. Before training, they generate samples of vertex neighborhoods via parameterized random walks. To represent asymmetric proximities, these techniques use two points for each vertex, one to represent the vertex as a source and the other to represent it as a destination. They are appropriate for link prediction, node labeling and community detection in social networks~\cite{node2vec}. These algorithms differ from our approach in important ways. First, they lose the physical interpretation since they use two points for each vertex in directed graphs. Second, they are semi-supervised algorithms that require labels or data about vertex similarities while our approach is an unsupervised approach that does not require any information other than the given directed graph. Third, our approach has a plug-and-play machine learning module that works well even with the Least Absolute Shrinkage and Selection Operator (LASSO) regression method, in which case it has a strongly polynomial runtime~\cite{wls}. HOPE~\cite{HOPE} is another algorithm that tries to preserve higher-order proximities, but it forgoes the use of random walks in favor of an approximate Singular Value Decomposition (SVD) of the similarity matrix. The top eigenvectors define the embedding space. Since HOPE relies on solving SVDs, it is prohibitively expensive for large graphs.

\section{Background}

\begin{figure}[!]
\centerfloat
\begin{subfigure}[b]{0.22\textwidth}
\centering
\includegraphics[width=\textwidth]{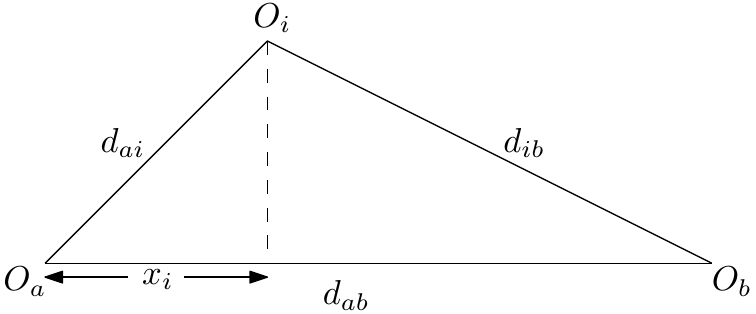}
\caption{}
\end{subfigure}
~
\begin{subfigure}[b]{0.22\textwidth}
\centering
\includegraphics[width=\textwidth]{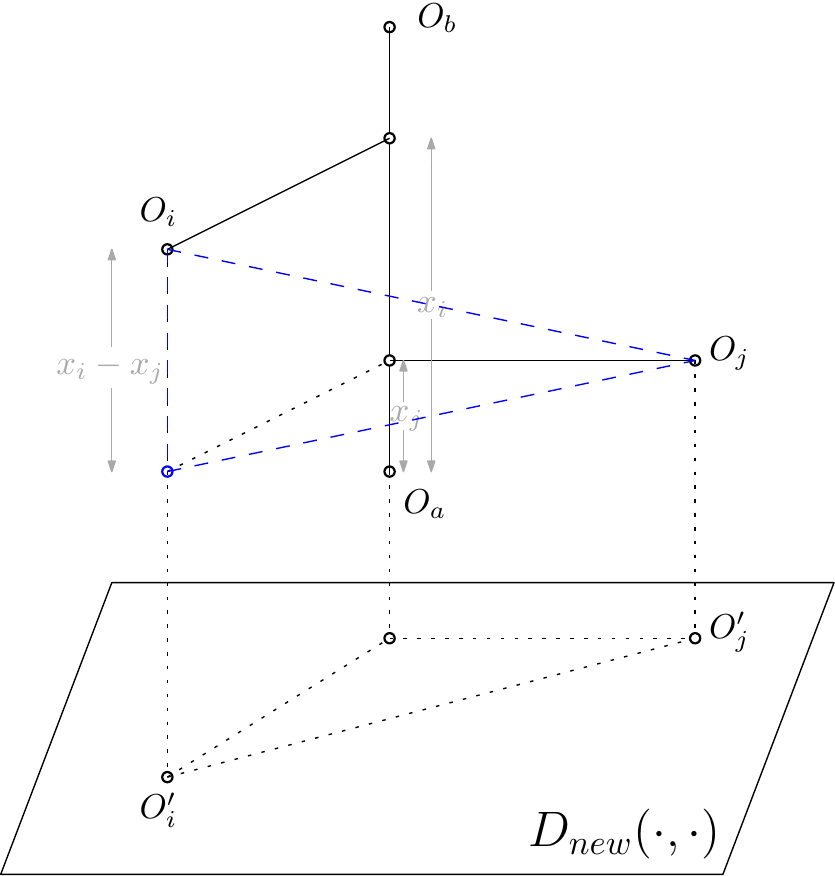}
\caption{}
\end{subfigure}
\caption{(a) shows the ``cosine law'' projection in a triangle. (b) illustrates how coordinates are computed and recursion is carried out in FastMap.}
\label{fig_fastmap_origins}
\end{figure}

\begin{figure*}[!]
\centerfloat
\begin{subfigure}[b]{0.45\textwidth}
\centering
\includegraphics[width=\textwidth]{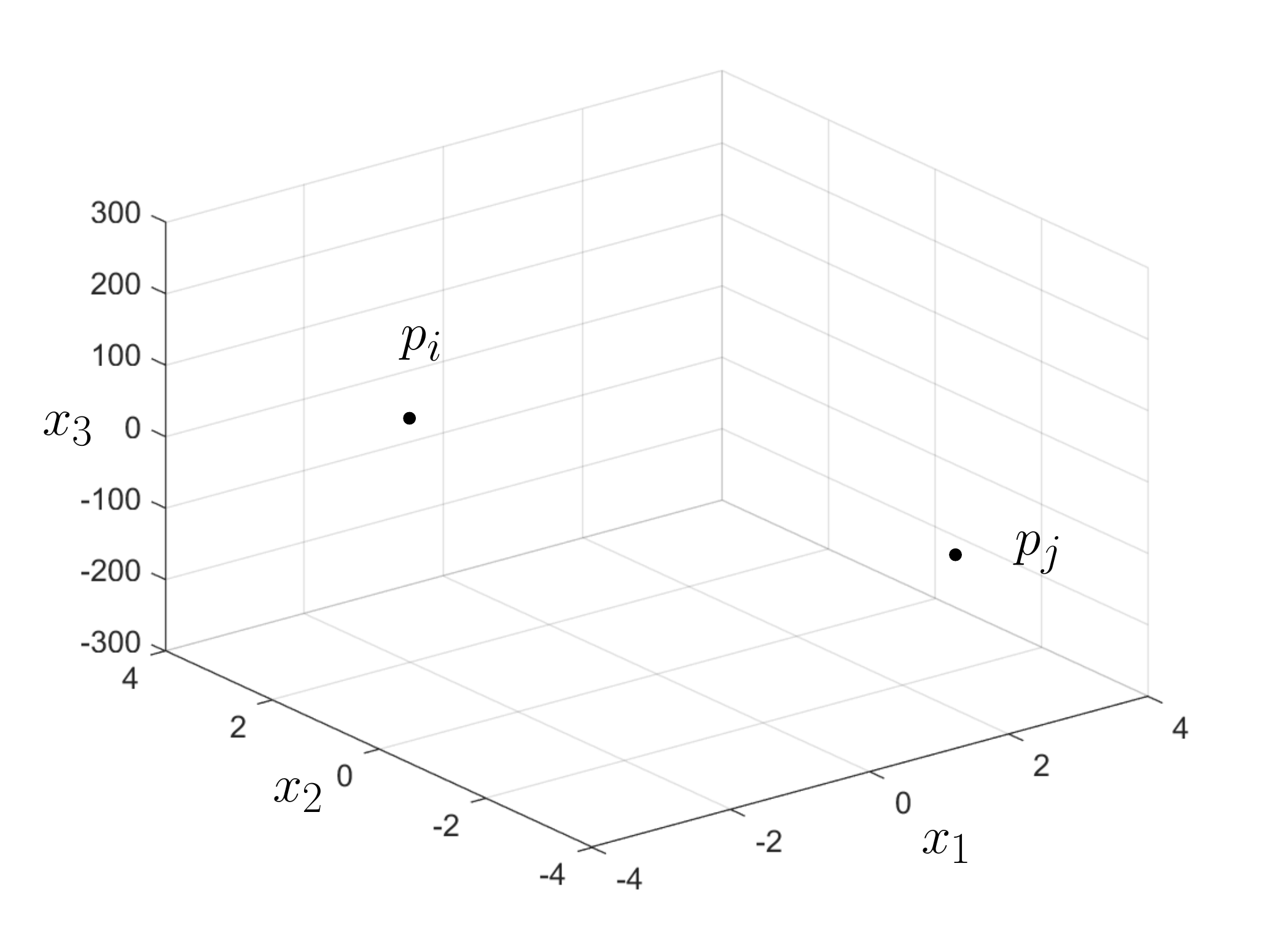}
\caption{Embedding produced by FastMap}
\end{subfigure}
~
\begin{subfigure}[b]{0.45\textwidth}
\centering
\includegraphics[width=\textwidth]{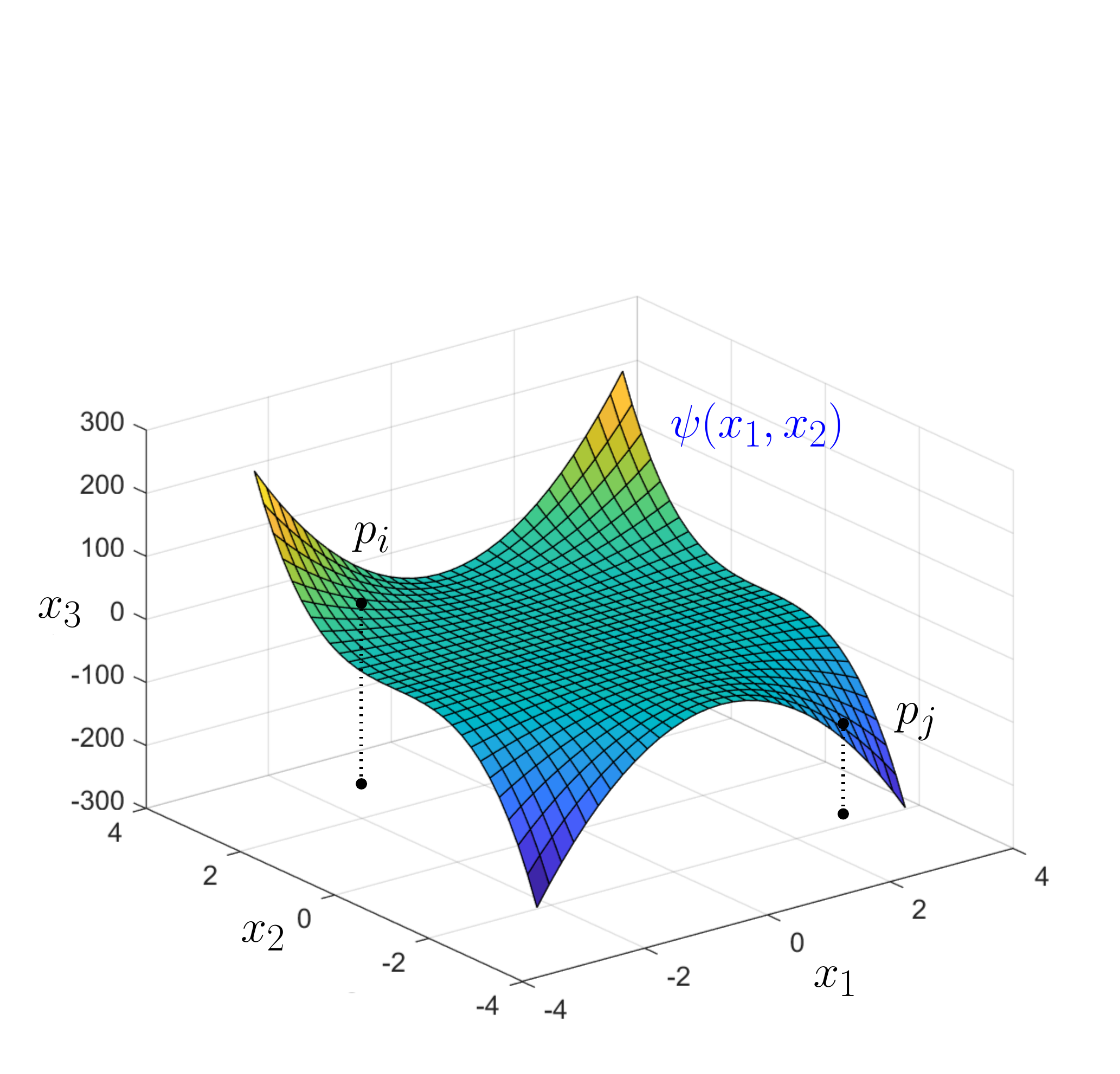}
\caption{Embedding produced by FastMap-D}
\end{subfigure}
\caption{Illustrates the difference between (a) the embedding produced by FastMap and (b) the embedding produced by FastMap-D. In (a), $3$ dimensions are used to represent the symmetric distances between vertices. In (b), $2$ dimensions are used to represent the symmetric average distances between vertices and the $3$rd dimension is used to represent the correction factors via the potential function $\psi$. Points $p_i$ and $p_j$ are the embeddings of vertices $v_i$ and $v_j$, respectively.}
\label{fig1_2}
\end{figure*}

FastMap~\cite{FL:SIGMOD:95} was introduced in the Data Mining community for automatically generating Euclidean embeddings of abstract objects. For example, if we are given objects in the form of long DNA strings, multimedia datasets such as voice excerpts and images or medical datasets such as ECGs or MRIs, there is no geometric space in which these objects can be naturally visualized. However, there is often a well-defined distance function between each pair of objects. For example, the \emph{edit distance}\footnote{The edit distance between two strings is the minimum number of insertions, deletions or substitutions that are needed to transform one to the other.} between two DNA strings is well defined although an individual DNA string cannot be conceptualized in geometric space. Clustering techniques, such as the $k$-means algorithm~\cite{A:BOOK:10}, are well studied in Machine Learning but cannot be applied directly to domains with abstract objects because they assume that objects are described as points in geometric space. FastMap revives their applicability by first creating a Euclidean embedding for the abstract objects that approximately preserves the pairwise distances between them.

In the Data Mining community, FastMap gets as input a \emph{complete} non-negative edge-weighted undirected graph $G=(V,E,w)$. Each vertex $v_i \in V$ represents an abstract object $O_i$. Between any two vertices $v_i$ and $v_j$, there is an edge $(v_i,v_j) \in E$ with weight $D(O_i,O_j)$ that corresponds to the symmetric distance between objects $O_i$ and $O_j$. A Euclidean embedding assigns a $K$-dimensional point $p_i \in \mathbb{R}^K$ to each object $O_i$. A good Euclidean embedding is one in which the Euclidean distance between any two points $p_i$ and $p_j$ closely approximates $D(O_i,O_j)$.

FastMap creates a Euclidean embedding in linear time by first assuming the existence of a very high dimensional embedding and then carrying out dimensionality reduction to a user-specified number of dimensions. In principle, it works as follows: In the first iteration, it heuristically identifies the farthest pair of objects $O_a$ and $O_b$ in linear time. It does this by initially choosing a random object $O_b$ and then choosing $O_a$ to be the object farthest away from $O_b$. It then reassigns $O_b$ to be the object farthest away from $O_a$. Once $O_a$ and $O_b$ are determined, every other object $O_i$ defines a triangle with sides of lengths $d_{ai}=D(O_a,O_i)$, $d_{ab}=D(O_a,O_b)$ and $d_{ib}=D(O_i,O_b)$. Figure~\ref{fig_fastmap_origins}(a) shows this triangle. The sides of the triangle define its entire geometry, and the projection of $O_i$ onto $\overline{O_a O_b}$ is given by $x_i = (d_{ai}^2 + d_{ab}^2 - d_{ib}^2) / (2d_{ab})$. FastMap sets the first coordinate of $p_i$, the embedding of object $O_i$, to $x_i$. In particular, the first coordinate of $p_a$ is $0$ and of $p_b$ is $d_{ab}$. Computing the first coordinates of all objects takes only linear time since the distance between any two objects $O_i$ and $O_j$ for $i,j \notin \{a,b\}$ is never computed.

In the subsequent $K-1$ iterations, the same procedure is followed for computing the remaining $K-1$ coordinates of each object. However, the distance function is adapted for different iterations. For example, for the first iteration, the coordinates of $O_a$ and $O_b$ are $0$ and $d_{ab}$, respectively. Because these coordinates fully explain the true distance $d_{ab}$ between them, from the second iteration onward, the remaining coordinates of $p_a$ and $p_b$ should be identical. Intuitively, this means that the second iteration should mimic the first one on a hyperplane that is perpendicular to $\overline{O_a O_b}$. Figure~\ref{fig_fastmap_origins}(b) explains this intuition. Although the hyperplane is never constructed explicitly, its conceptualization implies that the distance function for the second iteration should be changed to: $D_{new}(O'_i,O'_j)^2 = D (O_i, O_j)^2 - (x_i - x_j)^2$. Here, $O'_i$ and $O'_j$ are the projections of $O_i$ and $O_j$, respectively, onto this hyperplane, and $D_{new}(\cdot,\cdot)$ is the new distance function.

\section{FastMap-D}

In this section, we present FastMap-D, a generalization of FastMap to directed graphs. We assume that the given directed graph, $G=(V,E,w)$, is strongly connected, that is, there exists a path from any vertex $v_i \in V$ to any other vertex $v_j \in V$. While FastMap produces an embedding of the vertices in a Euclidean space for undirected graphs, FastMap-D produces an embedding of the vertices in a potential field for directed graphs. A $K$-dimensional potential field is a function $\psi : \mathbb{R}^K \rightarrow \mathbb{R}$. The potential field is used to capture asymmetric distances that are inherent in directed graphs.

Figure~\ref{fig1_2} illustrates the difference between the embeddings created by FastMap and FastMap-D. FastMap creates a $K$-dimensional point $p_i=\langle [p_i]_1,\ldots,[p_i]_K\rangle$ for each vertex $v_i$, as shown in Figure~\ref{fig1_2}(a). Here, the Euclidean distance $\|p_j - p_i\|_2 = \sqrt{\sum_{k=1}^K ([p_j]_k - [p_i]_k)^2}$ approximates the graph-based distance $d_G(v_i,v_j)$.

FastMap-D also creates a $K$-dimensional point $p_i=\langle [p_i]_1,\ldots,[p_i]_K\rangle$ for each vertex $v_i$, as shown in Figure~\ref{fig1_2}(b). However, $[p_i]_K = \psi([p_i]_1,\ldots,[p_i]_{K-1})$ for some $(K-1)$-dimensional potential field $\psi$. The FastMap-D distance $\| p_j - p_i \|_{\odot}$, defined to be $\sqrt{\sum_{k=1}^{K-1} ([p_j]_k - [p_i]_k)^2} + [p_j]_K-[p_i]_K$, approximates $d_G(v_i,v_j)$. The first term, $\sqrt{\sum_{k=1}^{K-1} ([p_j]_k - [p_i]_k)^2}$, approximates the symmetric average distance $\bar{d}_G(v_i,v_j) = \frac{d_G(v_i, v_j) + d_G(v_j, v_i)}{2}$, and the second term, $[p_j]_K-[p_i]_K$, approximates the asymmetric correction component $d_G(v_i,v_j) - \bar{d}_G(v_i,v_j)$.

\subsection{Algorithm Description}

\begin{algorithm*}[!]

\caption{Shows the FastMap-D algorithm. $G = (V,E,w)$ is a non-negative edge-weighted directed graph; $K_{max}$ is the user-specified upper bound on the dimensionality; $\epsilon$ is a user-specified threshold; $K+1 \leq K_{max}$ is the dimensionality of the computed embedding; $p_i$ is the embedding of vertex $v_i \in V$.}
\label{alg:fastmapd}

\function{\FastMapD}{
\KwIn{$G = (V,E,w)$, $K_{max}$ and $\epsilon$.}
\KwOut{$K+1$ and $p_i \in \mathbb{R}^{K+1}$ for all $v_i \in V$.}

\sub{Embed average distances in Euclidean space using FastMap.} {
${\tt pivots} \leftarrow \emptyset$\;
\For{$K = \{1, \ldots, K_{max}-1$\}}{
	\sub{Heuristically choose the farthest pair.} {
	Choose $v_a \in V$ uniformly at random and let $v_b \leftarrow v_a$\;
	\For(\tcp*[h]{$C$ is a small constant}){t = 1, \ldots, C}{
		\{$d_{ai}\}_{v_i \in V} \leftarrow$ \AvgToFroDist$(G,v_a)$\;
		$v_c \leftarrow \arg\max_{v_i} \{ d_{ai}^2 - \sum_{k=1}^{K-1} ([p_i]_k - [p_a]_k)^2 \}$\;
		\If{$v_c = v_b$}{
			Break\;
		}
		\Else{
			$v_b \leftarrow v_a$\;
			$v_a \leftarrow v_c$\;
		}
	}
	}
	${\tt pivots} \leftarrow {\tt pivots} \cup \{ v_a, v_b \}$\;
	\sub{Compute the $K$th coordinate.} {
	\{$d_{ai}\}_{v_i \in V} \leftarrow$ \AvgToFroDist$(G,v_a)$\;
	\{$d_{ib}\}_{v_i \in V} \leftarrow$ \AvgToFroDist$(G,v_b)$\;
	$d'_{ab} \leftarrow d_{ab}^2 - \sum_{k=1}^{K-1} ([p_b]_k - [p_a]_k)^2$\;
	\If{$d'_{ab} < \epsilon$} {
		Break\;
	}
	\For{each $v_i \in V$}{
		$d'_{ai} \leftarrow d_{ai}^2 - \sum_{k=1}^{K-1} ([p_i]_k - [p_a]_k)^2$\;
		$d'_{ib} \leftarrow d_{ib}^2 - \sum_{k=1}^{K-1} ([p_b]_k - [p_i]_k)^2$\;
		$[p_i]_K \leftarrow (d'_{ai} + d'_{ab} - d'_{ib}) / (2 \sqrt{d'_{ab}})$\;
	}
	}
}
}

\sub{Learn potential function and compute the last coordinate.} {
Let $S_1$ and $S_2$ be the two qualifying sets that define the sampling procedure\;
Let $\psi(x_1, \ldots, x_K) = \sum_{r=1}^M c_r x_1^{d_1^r} \ldots x_K^{d_K^r}$ be a multi-variate polynomial of degree $D$ with unknown coefficients $c_1, \ldots, c_M$; note that $M = \sum_{i=0}^D { {i+K-1} \choose {K-1} }$\;
Let $A = [A]_{ij}$ be a matrix of dimensions $|S_1||S_2| \times M$\;
Let $b = [b]_i$ be a vector of length $|S_1||S_2|$\;
Let $c = [c]_i$ be a vector of length $M$\;
\For{each $(v_i, v_j)$ such that $v_i \in S_1$, $v_j \in S_2$}{
	Let $1 \leq s \leq |S_1||S_2|$ be the current sampling index\;
	$[b]_s \leftarrow d_G(v_i,v_j) - \bar{d}_G(v_i,v_j)$\;
	Let $[A]_{sh}$ be the coefficient of $[c]_h$ in $\psi([p_j]_1, \ldots, [p_j]_K) - \psi([p_i]_1, \ldots, [p_i]_K)$ for $1 \leq h \leq M$\;
}
$c^* \leftarrow$ LASSO solution to $(Ac-b)^T (Ac-b)$\;
\For{each $v_i \in V$}{
	$[p_i]_{K+1} \leftarrow \psi([p_i]_1, \ldots, [p_i]_K)$\;
}
}
}

\function{\AvgToFroDist}{
\KwIn{$G = \langle V, E, w \rangle$ and a root $v_i \in V$.}
\KwOut{Average distance $\frac{d_G(v_i, v_j) + d_G(v_j, v_i)}{2}$ for all $v_j \in V$.}
Compute the shortest path tree rooted at $v_i$ in $G$ to get $d_G(v_i, v_j)$ for all $v_j \in V$\;
Let $G_R$ be $G$ with every edge reversed\;
Compute the shortest path tree rooted at $v_i$ in $G_R$ to get $d_{G_R}(v_i, v_j)$ (that is, $d_G(v_j, v_i)$) for all $v_j \in V$\;
\Return{$\frac{d_G(v_i, v_j) + d_G(v_j, v_i)}{2}$ for all $v_j \in V$\;}
}

\end{algorithm*}

Algorithm~\ref{alg:fastmapd} presents FastMap-D for directed graphs. The input is a non-negative edge-weighted directed graph $G = (V,E,w)$ along with two user-specified parameters $K_{max}$ and $\epsilon$. $K_{max}$ is the maximum number of dimensions allowed in the embedding. It bounds the amount of memory needed to store the embedding of any vertex. $\epsilon$ is the threshold that marks a point of diminishing returns when the distance between the farthest pair of vertices becomes negligible. The output is an embedding $p_i \in \mathbb{R}^{K+1}$ (with $K+1 \leq K_{max}$) for each vertex $v_i \in V$.

FastMap-D first embeds all vertices using average distances in a $K$-dimensional Euclidean space (lines 2-25). It then learns a potential function that is used to determine the $(K+1)$th coordinate (lines 26-38), which captures asymmetric distances as mentioned above.

\subsubsection{Embedding Average Distances:}

This phase of the algorithm (lines 2-25) is similar to the regular FastMap procedure that is applicable to undirected graphs. However, the input here is a directed graph, and the distances are asymmetric. Thus, we use the average distances $\bar{d}_G(v_i,v_j)$ as a symmetric measure derived from the directed graph. All pairwise distances or average distances are never explicitly computed since doing so would be computationally expensive. Instead, we use the function {\sc Average-Distance} that is invoked only $O(K_{max})$ times.

The function {\sc Average-Distance} (lines 39-43) computes $\bar{d}_G(v_i,v_j)$ for a given $v_i$ and all $v_j \in V$. It does this efficiently by computing two shortest path trees rooted at $v_i$. The first is computed on $G$ to yield $d_G(v_i,v_j)$ for all $v_j \in V$. The second is computed on $G_R$, which is identical to $G$ but with all edges reversed, to yield $d_{G_R}(v_i,v_j) = d_G(v_j,v_i)$ for all $v_j \in V$.

In each iteration of $K$ (line 4), the farthest pair of vertices ($v_a,v_b$) is heuristically chosen in near-linear time (lines 5-14). This pair of vertices is identified with respect to the residual distances for that iteration (line 9).\footnote{Note that $d_{ij} = \bar{d}_G(v_i,v_j)$.} The square of the residual distances in iteration $K$, $d_{ij}^2 - \sum_{k=1}^{K-1} ([p_j]_k - [p_i]_k)^2$, is the square of the original average distances minus the square of the Euclidean distances already explained by the first $K-1$ coordinates created so far. This is similar to the residual distances used in the $L_2$-variant of FastMap~\cite{jiaoyang2018}. The farthest pair of vertices, $v_a$ and $v_b$, are added to {\tt pivots}, a list of pivots, before the $K$th coordinate for each vertex is computed (lines 16-25) as follows: First, the {\sc Average-Distance} function is called on $v_a$ and $v_b$ to yield $d_{ai}$ and $d_{ib}$ for all $v_i \in V$ (lines 17-18). Then, the residual distances are computed (line 19), and FastMap's triangle projection rule is used to compute the $K$th coordinate (lines 23-25).

\subsubsection{Learning a Potential Function:}

This phase of the algorithm (lines 26-38) constructs a potential function $\psi(x_1,\ldots,x_{K})$ to account for asymmetric distances. The value of $\psi([p_i]_1,\ldots,[p_i]_{K})$ is recorded in the last coordinate of the embedding $[p_i]_{K+1}$. In this phase, a sampling procedure accompanies a learning procedure to construct $\psi(x_1,\ldots,x_{K})$. As shown in the pseudocode, $\psi(x_1,\ldots,x_{K})$ can be in the form of a multi-variate polynomial of degree $D$ on $x_1,\ldots,x_K$. It can also be in the form of a Neural Network (NN), as attempted in the next section. Of course, any machine learning algorithm can be used in this phase, but we choose to illustrate the pseudocode of the algorithm using polynomial fitting to facilitate our later discussion. While fitting a multi-variate polynomial can itself be done in many ways, here, we use LASSO to find the unknown coefficients of the multi-variate polynomial. The complexity of LASSO depends on the number of training samples~\cite{ehjt04}.

The sampling procedure uses two qualifying sets $S_1$ and $S_2$ (line 27), and all pairs $(v_i,v_j)$ with $v_i \in S_1$ and $v_j \in S_2$ are used as training samples (line 32). The number of training samples is therefore $|S_1||S_2|$. Different variants of FastMap-D can be created by varying the choices of $S_1$ and $S_2$. To keep the learning procedure efficient, both $S_1$ and $S_2$ cannot simultaneously be large subsets of $V$. On the other hand, restricting $S_1$ and $S_2$ to significantly smaller subsets can negatively impact the accuracy of the learning procedure. Therefore, FastMap-D chooses $S_1$ and $S_2$ judiciously, sometimes using {\tt pivots} computed in the first phase of the algorithm.

Consider a multi-variate polynomial $\psi(x_1,\ldots,x_{K})$ of degree $D$ having the form $\sum_{r=1}^M c_r x_1^{d_1^r} \ldots x_K^{d_K^r}$, where $d_1^r, \ldots, d_K^r \geq 0$ and $d_1^r + \ldots + d_K^r \leq D$ for $1 \leq r \leq M$. The number of terms, $M$, in the multi-variate polynomial is given by $\sum_{i=0}^D { {i+K-1} \choose {K-1} }$. We construct $c \in \mathbb{R}^{M}$ to be the vector of unknown coefficients of $\psi(x_1,\ldots,x_{K})$. $A \in \mathbb{R}^{|S_1||S_2| \times M}$ is a matrix in which row $s$ corresponds to sample $s$. For sample $s=(v_i,v_j)$, $\psi([p_j]_1, \ldots, [p_j]_K) - \psi([p_i]_1, \ldots, [p_i]_K)$ evaluates to a linear combination of the unknown coefficients and is desired to be equal to the correction factor $d_G(v_i,v_j) - \bar{d}_G(v_i,v_j)$, which is held in $[b]_s$. Therefore, $[A]_{sh}$ is equal to the coefficient of $[c]_h$ in $\psi([p_j]_1, \ldots, [p_j]_K) - \psi([p_i]_1, \ldots, [p_i]_K)$.\footnote{$\psi(x_1,\ldots,x_{K})$ has unknown coefficients on line 35 and, thus, $\psi([p_j]_1, \ldots, [p_j]_K) - \psi([p_i]_1, \ldots, [p_i]_K)$ evaluates to a linear combination of the unknown coefficients. However, on line 38, the unknown coefficients have been determined and, thus, $\psi([p_i]_1, \ldots, [p_i]_K)$ evaluates to a real number.}

Like the Ordinary Least Squares (OLS) method, LASSO minimizes $(Ac-b)^T (Ac-b)$ in $O(|S_1||S_2|M^2 + M^3)$ time to determine the unknown coefficients $c_1,\ldots,c_M$ (line 36). However, it uses $L_1$ regularization to address the regularization issues of OLS~\cite{wls}.

\subsection{Time Complexity}

FastMap-D makes $O(K_{max})$ calls to {\sc Average-Distance}. The time complexity of {\sc Average-Distance} is $O(|E| + |V|\log|V|)$. Therefore, the time complexity of the first phase of FastMap-D is $O(K_{max} (|E|+|V|\log|V|))$. Since LASSO takes $O(|S_1||S_2|M^2 + M^3)$ time, the overall time complexity of FastMap-D is $O(K_{max} (|E|+|V|\log|V|) + |S_1||S_2|M^2 + M^3)$, which is linear in $K_{max}$, near-linear in the size of the graph, linear in the number of training samples and exponential in the degree of $\psi$. In the next section, we discuss how to keep $|S_1||S_2|$ low. We also keep the degree of $\psi$ to a low constant.

\section{Experiments}

In this section, we present experimental results that demonstrate the benefits of FastMap-D. We conduct three kinds of experiments: (1) Comparing the accuracy of the embedding produced by FastMap-D to that of FastMap; (2) Evaluating different combinations of parameter values, specifically, the number of dimensions\footnote{$K + 1$ in pseudocode of Algorithm~\ref{alg:fastmapd}} $K$ and the degree of the potential function $D$; and (3) Evaluating the effectiveness of NNs trained on the FastMap coordinates of the vertices over NNs trained directly on the grid coordinates of the vertices. All experiments were conducted and evaluated on a 3.4GHz Intel-Xeon CPU with 64GB RAM. All algorithms were implemented in Python.

In our experiments, we also use a few implementation-level enhancements of the pseudocode of Algorithm~\ref{alg:fastmapd}. First, to exercise more control over the number of dimensions $K$, and to experiment with larger values of it, we try to avoid the break condition on line 21. We recognize that, if $d_{ab}^2 - \sum_{k=1}^{K-1} ([p_b]_k - [p_a]_k)^2 < \epsilon$ on line 11, the break condition on line 21 is satisfied. Therefore, we modify line 11 to reassign $v_a$ and $v_b$ randomly and continue the loop without breaking if indeed $d_{ab}^2 - \sum_{k=1}^{K-1} ([p_b]_k - [p_a]_k)^2 < \epsilon$. We also modify line 21 to set $d'_{ab}$ to $1$ instead of breaking the loop.\footnote{Otherwise, $d'_{ab}$ has a very low value, and the division in line 25 leads to numerical instability.} Second, to avoid obtuse triangles for the cosine law projection in Figure~\ref{fig_fastmap_origins}(a), we modify lines 23 and 24 so that $d'_{ai} \leftarrow \max(0, d_{ai}^2 - \sum_{k=1}^{K-1} ([p_i]_k - [p_a]_k)^2)$ and $d'_{ib} \leftarrow \max(0, d_{ib}^2 - \sum_{k=1}^{K-1} ([p_b]_k - [p_i]_k)^2)$.

Since the machine learning module of FastMap-D is designed to be a plug-and-play component, we implemented it using LASSO as well as an NN method.\footnote{Unless specified otherwise, FastMap-D refers to the version with LASSO.} For LASSO, we found it beneficial to set $S_1$ to {\tt pivots} since the pivots can be thought of as critical vertices identified in the first phase of FastMap-D. $S_2$ is set to a randomly chosen subset of vertices such that $|S_1||S_2| \ge M$. For training NNs, however, we generated training samples slightly differently (as described later in that subsection).

Although there exist benchmark instances for directed graphs, none of them come with the assurance of being strongly connected. For this reason, and to allow for a direct comparison with FastMap, the maps in this section are taken from a standard benchmark repository for undirected graphs~\cite{S:HOG:12}, which were also used in~\cite{cohen2018fastmap,jiaoyang2018}. For each map, we converted every edge into two directed edges in opposite directions to generate a directed version of it. We first created a virtual height $h(v)$ for each vertex $v \in V$ based on its 2D grid coordinates $(x_v,y_v)$. The height is assigned according to two possibilities: (a) polynomial function $h(v) = x_v + y_v^2 + (x_v + y_v)^3$, or (b) exponential function $h(v) = 1.01^{x_v} + 1.02^{y_v} + 1.03^{x_v + y_v}$. Then, we set $w(v_i,v_j)$ to be $2(h(v_j)-h(v_i))$ if $h(v_j) \geq h(v_i)$, and $(h(v_i)-h(v_j))/2$ otherwise.

To measure distortion, we use the Normalized Root Mean Square Error (NRMSE). We sample $N$ random distances between nodes, and compare them against their corresponding embedding distances $\|p_j - p_i\|_\odot$. To normalize the data coming from graphs of different sizes, the NRMSE is given by $\sigma / \bar{d}$ where
$\sigma=\sqrt{\frac{\sum_{1 \leq i \neq j \leq S}(d_{ij}-\|p_j - p_i\|_\odot)^2}{N}}$ and $\bar{d} = \frac{\sum_{1 \leq i \neq j \leq S}d_{ij}}{N}$.

\subsubsection{FastMap vs FastMap-D:}

\begin{figure*}[!]
\centerfloat
\begin{subfigure}[b]{0.45\textwidth}
\centering
\includegraphics[width=\textwidth]{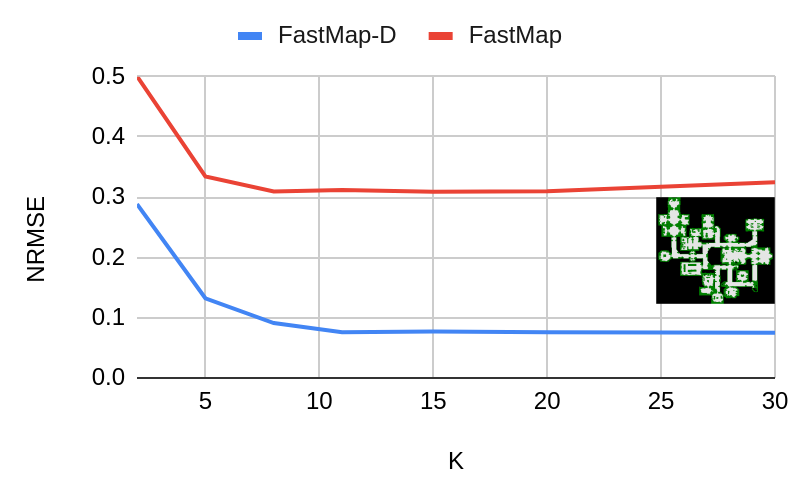}
\caption{hrt201n with polynomial height function}
\end{subfigure}
~
\begin{subfigure}[b]{0.45\textwidth}
\centering
\includegraphics[width=\textwidth]{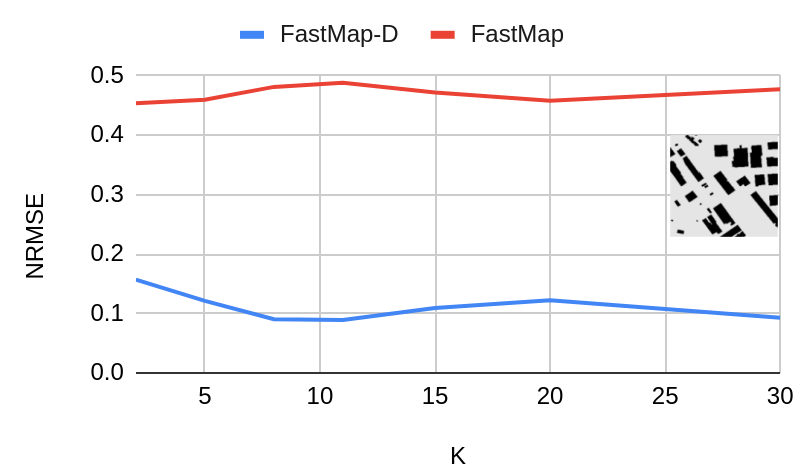}
\caption{Boston 2\_256 with polynomial height function}
\end{subfigure}

\begin{subfigure}[b]{0.45\textwidth}
\centering
\includegraphics[width=\textwidth]{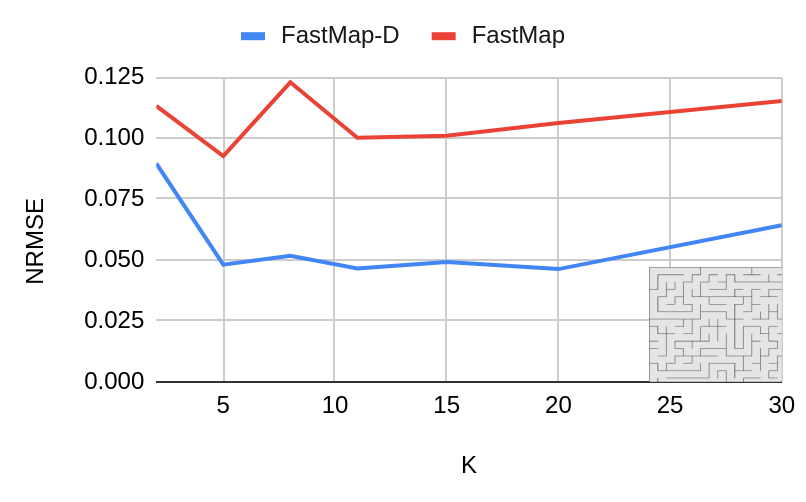}
\caption{maze 512-32-0 with polynomial height function}
\end{subfigure}
~
\begin{subfigure}[b]{0.45\textwidth}
\centering
\includegraphics[width=\textwidth]{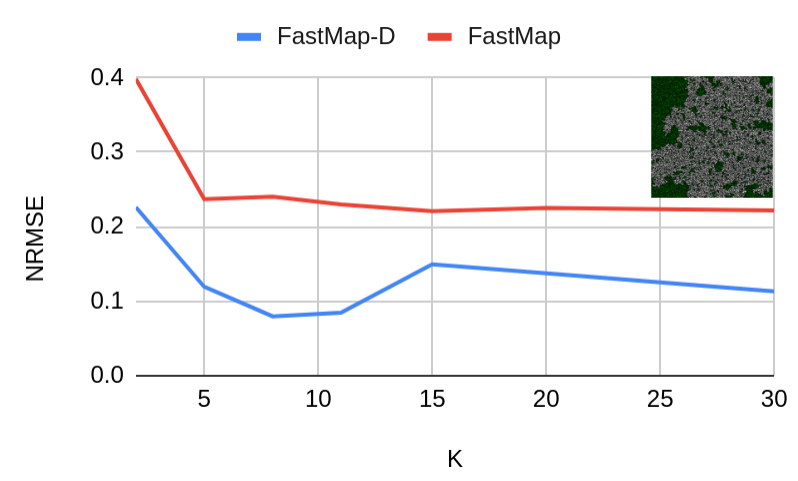}
\caption{random 512-40-0 with polynomial height function}
\end{subfigure}

\begin{subfigure}[b]{0.45\textwidth}
\centering
\includegraphics[width=\textwidth]{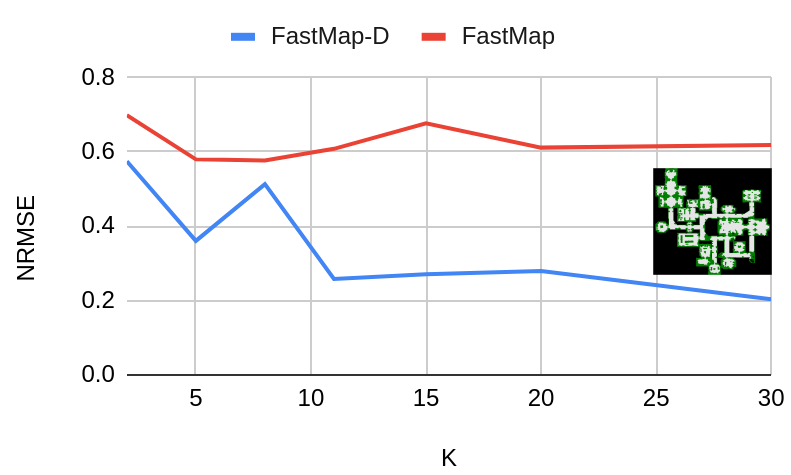}
\caption{hrt201n with exponential height function}
\end{subfigure}
~
\begin{subfigure}[b]{0.45\textwidth}
\centering
\includegraphics[width=\textwidth]{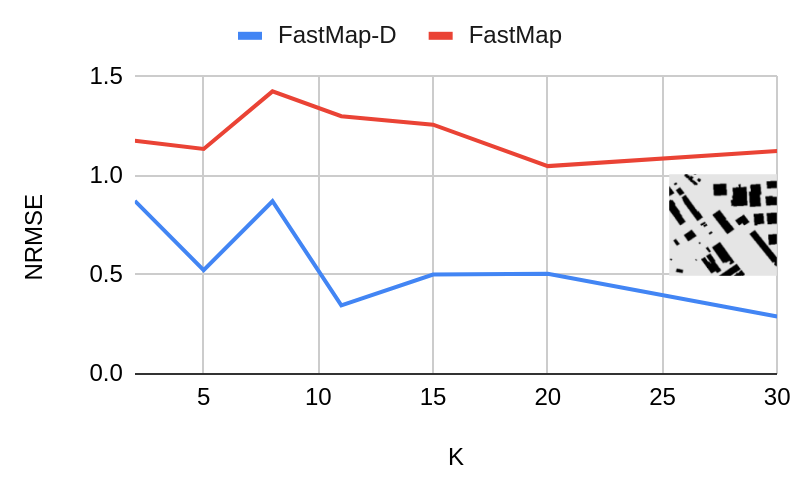}
\caption{Boston 2\_256 with exponential height function}
\end{subfigure}

\begin{subfigure}[b]{0.45\textwidth}
\centering
\includegraphics[width=\textwidth]{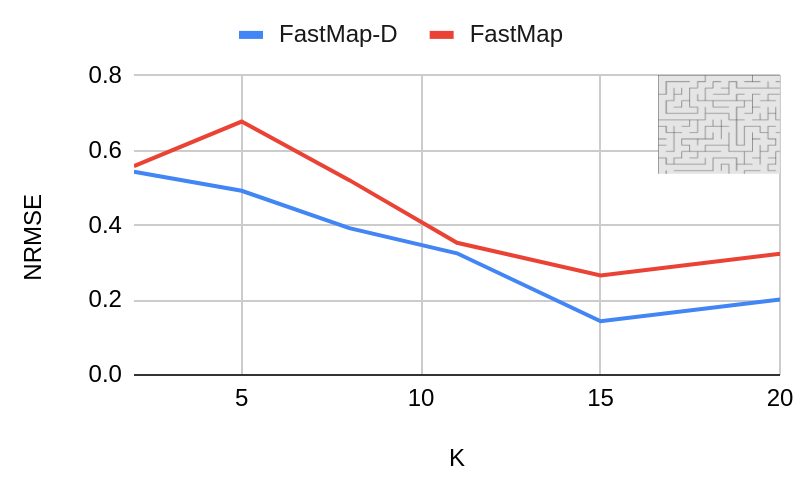}
\caption{maze 512-32-0 with exponential height function}
\end{subfigure}
~
\begin{subfigure}[b]{0.45\textwidth}
\centering
\includegraphics[width=\textwidth]{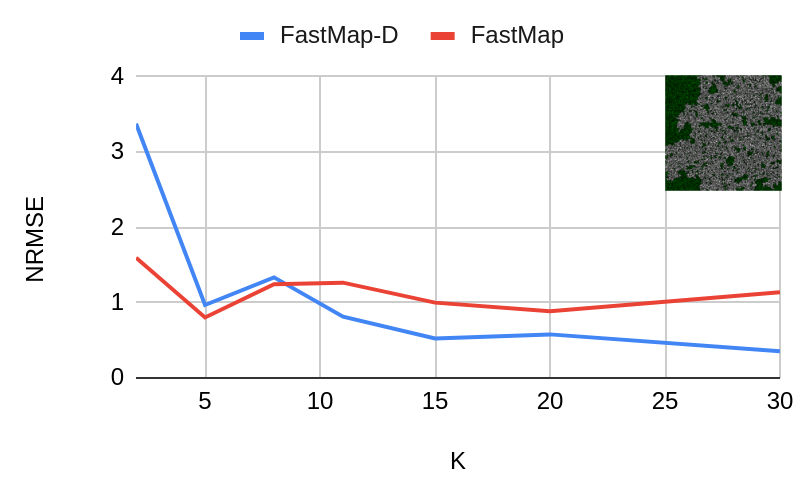}
\caption{random 512-40-0 with exponential height function}
\end{subfigure}

\caption{Shows the NRMSE values of FastMap and FastMap-D for different values of the number of dimensions $K$. In all cases, the degree $D$ of $\psi$ is $2$. The undirected version of the map is shown as an inlay.}
\label{fastmapd_vs_fastmap}
\end{figure*}

Figure~\ref{fastmapd_vs_fastmap} shows the NRMSE values of FastMap and FastMap-D for different values of the number of dimensions $K$ on four different kinds of maps. The top four panels show the results using the polynomial height function,\footnote{Using the polynomial height function for edge weights does not mean that the shortest path distances between vertices follow the same pattern. This is so because the map still has obstacles and the edge weights combine in complex ways to form shortest paths and graph distances.} and the bottom four panels show the results using the exponential height function. In all cases, we set the degree $D$ of $\psi$ to $2$. Since FastMap works only for undirected graphs, it can only embed the symmetric distances $\bar{d}_G(v_i,v_j)$. In other words, FastMap and FastMap-D differ only in the last coordinate that FastMap uses as an additional coordinate and FastMap-D uses as a correction factor to account for asymmetric distances.

We observe that FastMap-D outperforms FastMap on all maps for a sufficiently large number of dimensions $K$. Not only does FastMap-D outperform FastMap on mazes and random maps, but it also significantly outperforms FastMap on structured game maps and real-world city maps such as `hrt201n' and `Boston 2\_256'. We also note that the FastMap-D NRMSE values often decrease faster than the FastMap NRMSE values for increasing $K$. This shows that FastMap-D utilizes additional dimensions better than FastMap does.

\subsubsection{Varying FastMap-D Parameter Values:}

\begin{figure*}[!t]
\centerfloat
\begin{subfigure}[b]{0.45\textwidth}
\centering
\includegraphics[width=\textwidth]{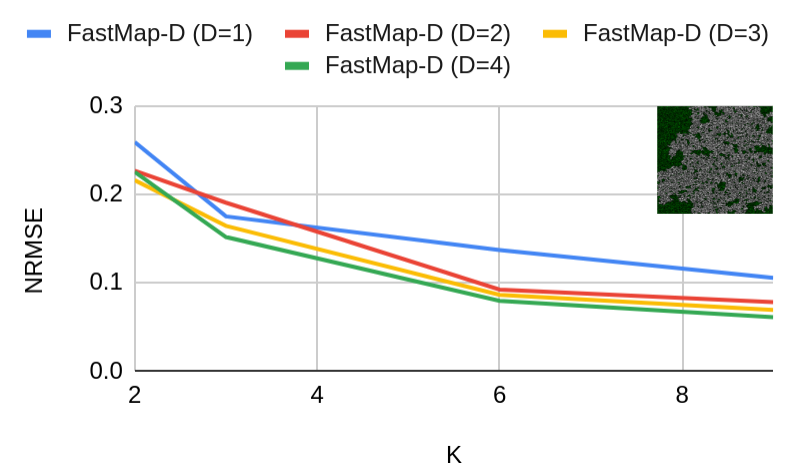}
\caption{random 512-40-0 with polynomial height function}
\end{subfigure}
~
\begin{subfigure}[b]{0.45\textwidth}
\centering
\includegraphics[width=\textwidth]{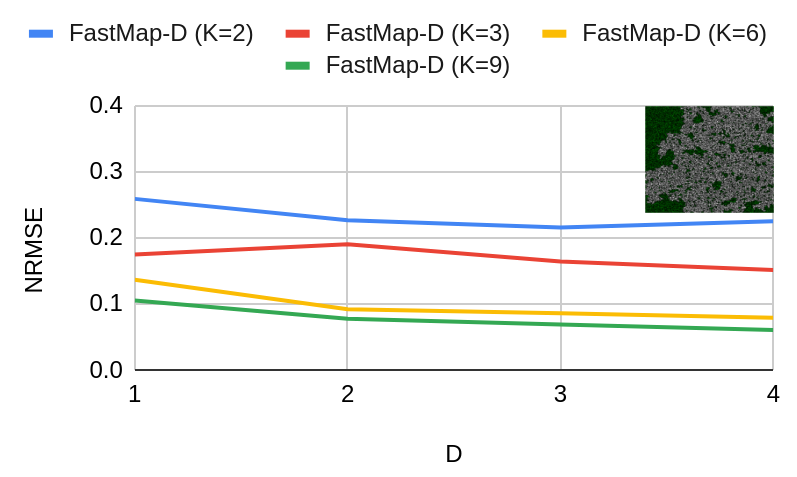}
\caption{random 512-40-0 with polynomial height function}
\end{subfigure}
\caption{Shows the NRMSE values of FastMap-D with different parameter values. $K$ is the number of dimensions; and $D$ is the degree of $\psi$. The undirected version of the map is shown as an inlay.}
\label{fastmapd_params}
\end{figure*}

Figure~\ref{fastmapd_params} shows the effect of $K$ and $D$ on the NRMSE values of FastMap-D for a representative map. In general, increasing $K$ improves the NRMSE values. However, increasing $D$ is not always very helpful.

\subsubsection{NNs on FastMap Coordinates:}

\begin{table}[t!]
\scriptsize
\centering
\begin{tabular}{|c|c|c|c|}
\hline
Instance & Direct NN & FastMap-D (NN) & FastMap-D (LASSO) \\
\hline
Lak503d Poly & 1.238 & 0.048 & 0.042 \\
Lak503d Exp & 0.901 & 0.071 & 0.089 \\
hrt201n Poly & 0.944 & 0.028 & 0.077 \\
hrt201n Exp & 1.239 & 0.083 & 0.271 \\
Boston 2\_256 Poly & 0.994 & 0.039 & 0.109 \\
Boston 2\_256 Exp & 2.795 & 0.043 & 0.501 \\
\hline
\end{tabular}
\vspace{1em}
\caption{Compares the NRMSE values of the direct NN approach and FastMap-D on a few representative maps. The best NN designed for the direct approach uses $4$, $1000$, $500$, $200$, $200$ and $1$ nodes in fully connected consecutive layers. The NN designed for FastMap-D uses $K = 15$ and $30$, $1000$, $500$ and $1$ nodes in fully connected consecutive layers. $538560$, $709530$ and $1458360$ training samples were used for the Lak503d, hrt201n and Boston 2\_256 maps, respectively. `Poly' and `Exp' indicate polynomial and exponential height functions, respectively.}
\label{tab:NNs}
\end{table}

NNs can learn pairwise distances between the vertices in a grid map. Naively applied, an NN can be trained on the 2D grid coordinates of the source and destination vertices. However, there are several problems with this direct approach. First, it is not applicable to general graphs where vertices do not have coordinates. Second, even for grid maps, the feature set is very small since it is limited to the grid coordinates. Third, it remains oblivious of many parts of the graph even with a large number of training samples since there are a quadratic number of pairs of source and destination vertices.

There are several benefits of using the FastMap coordinates instead of the grid coordinates for training NNs. First, this approach is applicable to general graphs. Second, the feature set is larger and depends on the user-controlled parameter $K$. Third, since the link structure of the graph is summarized in the FastMap coordinates, not too many samples are required. In fact, we simply use $|{\tt pivots}|$ randomly selected vertices, compute the shortest path trees rooted at each of them, and draw training samples only from these trees with the source vertex restricted to be the root vertex. This keeps the number of training samples linear in the size of the graph.

Table~\ref{tab:NNs} shows the benefit of training NNs on the FastMap coordinates compared to training them on the grid coordinates. For the same number of training samples, the NRMSE values of FastMap-D with NNs are significantly smaller than those of the direct approach with NNs that uses the grid coordinates. They are also smaller than those of FastMap-D with LASSO.

\section{Conclusions and Future Work}

In this paper, we generalized FastMap for undirected graphs to FastMap-D for directed graphs. FastMap-D efficiently embeds the vertices of a given directed graph in a potential field. Unlike a Euclidean embedding, a potential-field embedding can represent asymmetric distances. FastMap-D uses a machine learning module to learn a potential function that defines the potential field. In experiments, we demonstrated the advantage of FastMap-D on various kinds of directed graphs. An important upshot of our approach is that applying machine learning algorithms to the FastMap coordinates of the vertices of a graph is much better than applying them directly to the grid coordinates of the vertices since the FastMap coordinates capture important information of the link structure of the graph - not to mention that the grid coordinates are not even defined for general graphs.

In future work, we will apply FastMap-D to very large directed graphs, such as knowledge graphs, and to intensional graphs, such as in automated planning and plan visualization. The success of FastMap-D exemplifies the benefits of using the FastMap coordinates as features for machine learning algorithms, and we hope to do the same for other graph problems that involve machine learning.

\section{Acknowledgements}

The research at Arizona State University was supported by ONR grants N00014-16-1-2892, N00014-18-1-2442, N00014-18-1-2840 and N00014-19-1-2119. The research at the University of Southern California was supported by NSF grants 1724392, 1409987, 1817189, 1837779 and 1935712.

\bibliography{references}
\bibliographystyle{aaai}
\end{document}